\documentclass[sigconf]{acmart}

\AtBeginDocument{%
  }

\setcopyright{none}
\copyrightyear{2027}
\acmYear{2027}
\acmConference[KDD '27]{The 33rd ACM SIGKDD Conference on Knowledge Discovery and Data Mining}{August 1--5, 2027}{San Jose, CA, USA}
\acmDOI{}
\acmISBN{}

\usepackage{algorithm}
\usepackage{algorithmic}
\usepackage{booktabs}
\usepackage{multirow}
\usepackage{placeins}
\usepackage{enumitem}
\usepackage{listings}
\newcommand{\tblhead}[1]{\textbf{#1}}
\newcommand{\tblbest}[1]{\textbf{#1}}
\newcommand{\tblsecond}[1]{\underline{#1}}
\newcommand{\tabledash}{\textemdash}

\begin{document}

\title{MAS4BE: An LLM-Driven Multi-Agent Sandbox for B2B Ecosystem Simulation}

\author{Zeyu Wang}
\authornote{Both authors contributed equally to this research.}
\affiliation{%
  \institution{Zhejiang University}
  \city{Hangzhou}
  \country{China}}
\additionalaffiliation{%
  \institution{Alibaba Group}
  \city{Hangzhou}
  \country{China}}
\email{zywang03@zju.edu.cn}

\author{Yao Zhao}
\authornotemark[1]
\authornote{Corresponding authors.}
\affiliation{%
  \institution{Alibaba Group}
  \city{Hangzhou}
  \country{China}}
\email{zhaoyao.zhao@taobao.com}

\author{Tengyue Ma}
\authornotemark[3]
\affiliation{%
  \institution{Alibaba Group}
  \city{Hangzhou}
  \country{China}}
\email{xianyue.mty@taobao.com}

\renewcommand{\shortauthors}{Wang et al.}

\begin{abstract}
B2B e-commerce platforms sell services such as paid memberships and advertising to large merchant populations, making the allocation of limited sales resources a central planning problem. Existing approaches typically rank merchants by predicted purchase propensity to prioritize outreach. Although effective at merchant ranking, these models learn only static associations from historically observed sales actions; they struggle to capture the state evolution and interaction processes underlying purchase propensity, let alone simulate the outcomes of alternative resource allocations before execution.

To address these limitations, we present MAS4BE, an LLM-driven multi-agent sandbox that recasts B2B resource allocation as ecosystem simulation. It transforms longitudinal operational logs into an executable environment in which stateful merchant and sales agents co-evolve through repeated interactions, linking individual decisions to population dynamics. This design makes population trajectories traceable, enables interpretable, model-based what-if analysis of alternative resource allocations, and generates transferable population signals for downstream tasks.

MAS4BE has been deployed for daily decision support in a real B2B sales environment. Results from offline experiments and a randomized field experiment show that it faithfully reproduces daily conversion trajectories, responds coherently to contact-budget interventions, and generates population signals that improve the mean AUC-ROC of downstream prediction models. These results demonstrate MAS4BE's cross-product generalization, ability to explore resource allocations under controlled interventions, and downstream utility.
\end{abstract}



\maketitle

\section{Introduction}
\label{sec:intro}

B2B e-commerce platforms sell services such as paid memberships and advertising to large merchant populations~\cite{lilien2016b2b}. In the production setting we study, a sales team serves tens of thousands of merchants, yet each representative can only contact a small fraction of their portfolio on any given day. Planners must decide which merchants to reach out to and how to spread limited outreach across the team. A standard approach is to predict each merchant's purchase propensity from historical data and rank accordingly~\cite{dhaen2013prospect,chapelle2014simple}. This works well for the narrow question of who should be contacted first under the current workflow, but planning also needs to answer harder ones. How would the population evolve under a different allocation, and what happens when the contact budget changes? In practice each such question requires its own model, an uplift model for causal effects~\cite{radcliffe2011real,athey2016recursive} or a forecasting model for trajectories, each with its own data pipeline, assumptions, and upkeep. As products and scenarios multiply, these siloed pipelines become expensive to extend, and they cannot share state or adapt to plans they were not built for.

Instead of adding more models, we recast the problem as \emph{ecosystem simulation}. We present MAS4BE (Multi-Agent Sandbox for B2B Ecosystem), an LLM-driven sandbox in which stateful merchant and sales agents co-evolve through day-by-day, capacity-constrained interactions, connecting individual decisions to population dynamics. The architecture is hybrid, with LLMs handling context-sensitive simulation while deterministic code manages state transitions, guardrails, and fallbacks. It combines ideas from generative agents~\cite{park2023generative,gao2024large}, agent-based modeling~\cite{bass1969new,rand2011agent}, and digital twins~\cite{barricelli2019digital,tao2019digital,fuller2020digital} into a system that planners use daily to compare scenarios before finalizing outreach decisions (\S\ref{sec:system}).

From one simulation run, planning teams get three outputs. First, \textbf{population trajectories} with full event-level provenance show planners how daily conversion is expected to evolve. Second, \textbf{what-if replay} lets planners compare scenarios side by side from identical starting conditions without building a separate model. Third, \textbf{population signals} feed simulated dynamics into existing prediction models, boosting their accuracy.

MAS4BE has been deployed and validated through extensive offline experiments and a randomized online field experiment. Simulated trajectories closely track observed conversion, reaching Pearson $r$ up to $0.89$, and the deployed workflow yields a $31.2\%$ relative lift in first-purchase rate at $p=0.018$, with ablation confirming the gain traces to sandbox-derived signals.

To our knowledge, MAS4BE is among the first LLM-driven multi-agent simulations deployed for daily B2B sales allocation and validated through both offline replay and a randomized online experiment; our emphasis is on this combination of production deployment and dual validation, not on novelty alone. We make four contributions:
\begin{itemize}[nosep,leftmargin=1.2em]
\item \textbf{Deployed ecosystem simulation for B2B sales planning.} We design and deploy MAS4BE, a daily-running multi-agent simulation in which merchant and sales agents co-evolve under production constraints, serving as the core decision-support tool for sales planning (\S\ref{sec:problem}, \S\ref{sec:system}).
\item \textbf{Unified capabilities from a single simulation.} Trajectory forecasting, scenario comparison, and downstream signal generation all arise from one stateful execution; no per-task model is needed, and the sandbox complements the existing downstream ranker. Products introduced at evaluation time achieve high fidelity from the shared agent design and product context under pre-evaluation calibration alone, using only a product-level threshold and market-coefficient calibration, with no product-specific tuning or retraining of the shared framework (\S\ref{sec:system}, \S\ref{sec:eval}).
\item \textbf{Comprehensive offline and online validation.} We validate MAS4BE through four complementary protocols spanning trajectory fidelity, what-if coherence, signal transfer, and a randomized field experiment, demonstrating both offline accuracy and real-world business impact (\S\ref{sec:eval}).
\item \textbf{Operational design lessons.} Role-specific model assignment, temporal auditing of grouping features, and guardrail-based state management offer practical guidance for deploying LLM-agent systems in production. Although developed for B2B sales, the design principles are domain-agnostic and extend naturally to other resource-constrained allocation settings (\S\ref{sec:deployment_lessons}).
\end{itemize}

\section{Problem Setting}
\label{sec:problem}

A B2B platform sells paid services (products) through sales representatives. Each representative manages a merchant portfolio with limited daily contact capacity. Unlike propensity ranking, which scores merchants independently from a static snapshot, the sandbox advances the joint state of all agents day by day: each day's contacts, responses, and outcomes shape the next day's state. MAS4BE therefore evaluates how a population evolves under a specified plan; it does not predict individual purchase outcomes or search for optimal policies.

Let $y_{m,P,D}\in\{0,1\}$ indicate whether merchant~$m$ first purchases product~$P$ on day~$D$. The at-risk population $\mathcal{A}(P,D)$ consists of merchants who have not yet purchased~$P$ before day~$D$, and the daily conversion rate $r(P,D)$ is the fraction of at-risk merchants who purchase on that day. Target-day labels are evaluation data, not simulation inputs. A daily allocation~$u_D$ specifies per-representative contact capacities and a selection rule; each representative~$s$ selects contacts from at-risk merchants in their portfolio, subject to
\begin{equation}
\mathcal{C}_s(P,D;u_D) \subseteq \mathcal{A}(P,D) \cap \mathcal{M}_s(D),\quad |\mathcal{C}_s(P,D;u_D)| \leq K_s(D).
\label{eq:allocation_constraint}
\end{equation}
For horizon~$H$, $u{=}(u_D,\ldots,u_{D+H-1})$ is a scenario sequence.
Throughout, \emph{what-if} and \emph{counterfactual} refer to model-based comparisons between sandbox trajectories initialized from the same state, not identified causal effects.

The sandbox advances the joint state~$\mathbf{Z}_D$ of all agents one day at a time using only planning-time information: contacts are selected, sales agents act, merchant agents respond, and validated events update persistent state before day~$D{+}1$. Each step produces the population trajectory~$\widehat{\mathbf r}(u)$, an event trace~$\mathcal{E}(u)$, and downstream signals~$\mathbf{g}(u)$ (\S\ref{sec:output_interface}).

The framework is product-agnostic by design: agent architectures, state dimensions, population mechanisms, and transition logic are shared across products; only product descriptors, penetration rates, and calibration parameters provide product-specific context. To validate this generality, we deploy MAS4BE on four anonymized products (A--D) within the pilot business unit. Products~A and~B undergo 46~days of development-time profile evolution; Products~C and~D are introduced at evaluation time with the agent design, schema, prompts, and shared settings frozen from A/B, testing within-platform transfer of a single framework instance to new products. This is a cold-start setting in which target-product \emph{individual} training labels are unavailable and only pre-evaluation calibration, fit from product-level aggregate history, is used.

\section{System Design}
\label{sec:system}

\begin{figure*}[t]
  \centering
  \includegraphics[width=1\textwidth]{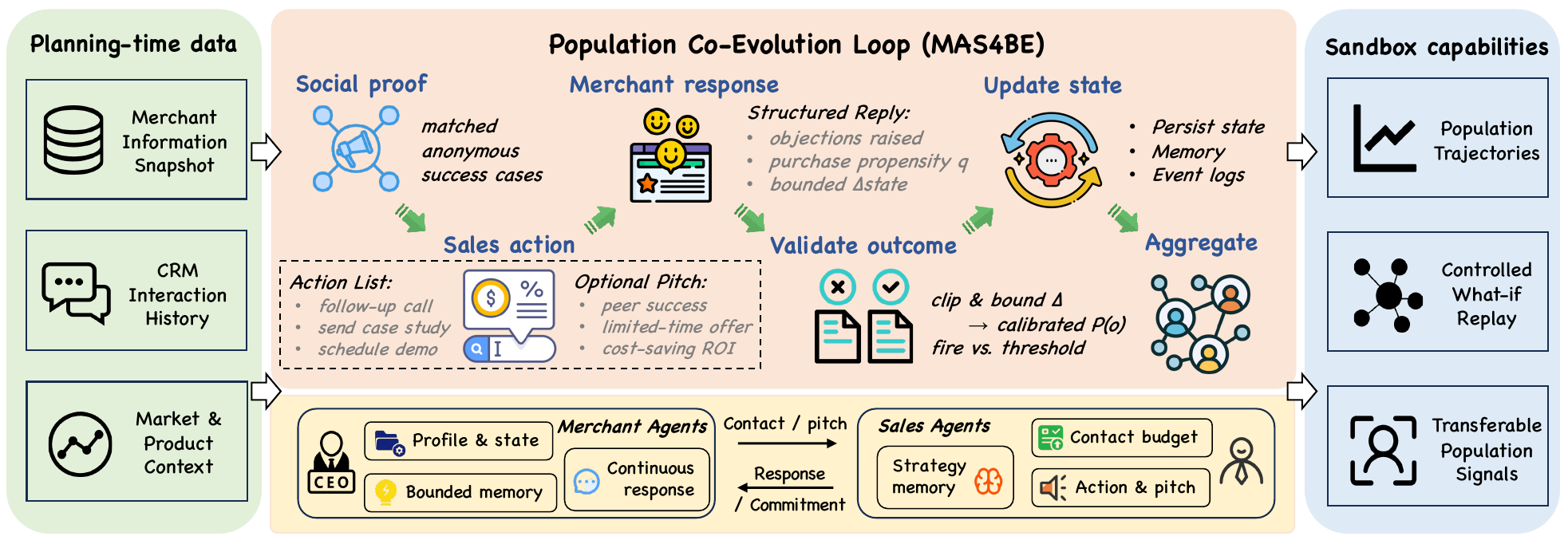}
  \caption{MAS4BE architecture and information boundary. Planning-time records initialize stateful merchant and sales agents; constrained daily interactions and population mechanisms produce trajectories, what-if replays, and transferable population signals. Target-day and future outcomes are hidden from all agent prompts; only lagged outcomes enter eligibility and aggregate context.}
  \Description{Three-part architecture diagram. Left: merchant snapshots, CRM histories, and market context feed a deterministic adapter. Center: sales and merchant agents interact under budget constraints, bounded state, population mechanisms, and event tracing, with target-day outcomes hidden. Right: three outputs---trajectories, what-if replays, and population signals.}
  \label{fig:architecture}
\end{figure*}

MAS4BE turns longitudinal records into a merchant--sales ecosystem that advances daily. Throughout, a \emph{merchant agent} denotes the in-sandbox simulated counterpart of a real-world \emph{merchant}: the entity, its data, and the unit of randomization. LLMs generate context-sensitive decisions; deterministic code validates, clips, advances state, and provides complete fallbacks. Figure~\ref{fig:architecture} maps the resulting architecture to its runtime stages and three outputs. We describe data initialization (\S\ref{sec:data_init}), the daily execution loop (\S\ref{sec:daily_exec}), hybrid interaction design (\S\ref{sec:interaction}), population dynamics (\S\ref{sec:pop_dynamics}), output interface (\S\ref{sec:output_interface}), and deployment safeguards (\S\ref{sec:deployment}).

The multi-agent topology mirrors the operational structure. Each sales agent is linked to a disjoint merchant portfolio, forming a bipartite assignment graph. Merchants in the same industry or region share aggregate context (peer purchase events, penetration trends, and market signals), so one merchant's outcome can shift the landscape for nearby agents. Sales agents share a collective strategy memory that propagates effective action--response patterns across the team without direct inter-agent communication. This layered connectivity (portfolio-constrained dyadic interactions (\S\ref{sec:interaction}), region-scoped population dynamics (\S\ref{sec:pop_dynamics}), and team-wide strategy diffusion) mirrors the real organizational topology. We adopt it as a design choice and do not claim empirical superiority over alternatives.

\subsection{Planning-Time Data and State Initialization}
\label{sec:data_init}

The deployed path joins three sources: daily merchant snapshots, sanitized CRM interaction histories, and first-purchase labels across all products in the pilot business unit. A deterministic adapter produces (i) observed profiles, behavioral states, and bounded memories; (ii) sales portfolios, capacities, and strategy memory; and (iii) industry, region, market, and product aggregates. Raw records remain outside the agent loop while assignment and group structure are preserved.

For a run on day $D$, the adapter uses the operational snapshot available at recommendation time. Transaction-derived eligibility and product penetration are truncated at $D{-}1$; $y_{m,P,D}$ and all future outcomes are hidden from both agent types and enter only the explicitly stated offline calibration and evaluation procedures. All runtime fields pass the same temporal cutoff and an allowlist; identifiers are hashed and CRM text is reduced to sanitized summaries.

\subsection{Daily Stateful Execution}
\label{sec:daily_exec}

Let $\mathbf{Z}_D$ contain all persistent merchant and sales states, memories, eligibility, assignments, and population statistics. One daily tick advances $\mathbf{Z}_{D+1}=T_{\Theta}(\mathbf{Z}_D,\mathbf{X}^{\mathrm{plan}}_D,u_D,\epsilon_D)$ under fixed settings $\Theta$ and seed-controlled draws $\epsilon_D$. Algorithm~\ref{alg:sandbox} specifies the order: profile and market refresh precede background dynamics; all contact sets are selected before any contact executes, preventing within-day ordering effects. Different representatives run concurrently on disjoint portfolios. Accepted transitions are appended to the event log before state carries to day $D{+}1$.

\begin{algorithm}[t]
\caption{One-day sandbox execution loop for a single product.}
\label{alg:sandbox}
\begin{algorithmic}[1]
\REQUIRE Product $P$, state $\mathbf{Z}_D$, planning data $\mathbf{X}^{\mathrm{plan}}_D$, scenario $u_D$
\STATE Refresh profiles and planning-time market context from $\mathbf{X}^{\mathrm{plan}}_D$
\STATE Apply decay, peer and market effects, and triggered autonomous updates
\STATE Select all feasible contact sets $\{\mathcal{C}_s(P,D;u_D)\}_{s\in\mathcal{S}}$
\FOR{each selected merchant--sales pair $(m,s)$}
  \STATE Retrieve eligible anonymous social proof and generate a sales action
  \STATE Generate and validate the merchant response and bounded state changes
  \STATE Update merchant and sales state; draw outcomes using $\epsilon_D$
  \STATE Append the action, response, transition, and outcome events
\ENDFOR
\STATE Synchronize aggregate sales-team patterns and recompute population summaries
\RETURN $\mathbf{Z}_{D+1}$, $\widehat{\mathbf r}(P,D;u_D)$, $\mathcal{E}(u_D)$, and $\mathbf{g}(u_D)$
\end{algorithmic}
\end{algorithm}

The event log preserves each execution as an ordered trace. For a paired comparison, fixing model, prompts, initial state, planning data, exogenous context, and seed holds conditions constant; only $u_D$ changes.

\subsection{Hybrid Merchant--Sales Interaction}
\label{sec:interaction}

Each daily contact pairs one sales agent with one merchant agent. The division of labor is strict: the LLM generates context-sensitive responses (objections, commitment signals, propensity estimates), while deterministic code validates every proposed state change, computes calibrated outcome probabilities, and provides complete fallback when LLM output is malformed or unavailable. The LLM therefore influences persistent state only through validated, bounded transition inputs.

\paragraph{Merchant agents.}
Each agent combines its observed profile, a behavioral state vector, and four bounded memory layers inspired by the layered memory architecture of generative agents~\cite{park2023generative}. The state spans readiness, beliefs and constraints, relationship history, and outcome risks; dimensions are included only when they can be initialized from planning-time data, changed by an interaction, and given an operational interpretation. For a contact, the merchant LLM sees allowlisted state, bounded memory, the sales action and pitch template, peer context, and market context. It returns a schema-constrained response with objections, commitment, propensity, and proposed state changes.

Because raw LLM outputs can be arbitrarily noisy, all state changes pass through a bounded update that constrains their magnitude. Each validated delta $\Delta_j\in[-0.5,0.5]$ is applied as $s_j\!\leftarrow\!\operatorname{clamp}(s_j+g_j(s_j,\Delta_j),\,0,\,1)$, where $g_j(s,\Delta)=\Delta(1{-}s)$ for $\Delta\!\ge\!0$ and $\lambda_j\,\Delta\,s$ otherwise. Positive changes thus saturate near the upper bound, while $\lambda_j\!>\!1$ amplifies negative shifts for dimensions like trust and relationship warmth. When the LLM returns propensity~$q$, purchase intent blends the rule-updated value with the LLM output: $s_{\rm intent}\!\leftarrow\!(1{-}\alpha)\,\widetilde{s}_{\rm intent}+\alpha\,q$. The weight $\alpha{=}0.7$ lets the LLM estimate dominate this single intent dimension, but its influence stays bounded: intent affects outcomes only through the calibrated function (Eq.~\ref{eq:outcome}), never firing events directly.

For outcome $o$, the runtime computes
\begin{equation}
  P(o)=\sigma\!\left(\kappa(z_o-\theta_o)\right)d_{\mathrm{fatigue}}d_{\mathrm{repeat}},
  \label{eq:outcome}
\end{equation}
where $z_o$ combines outcome-relevant state and decision-model likelihood, $\theta_o$ is the calibrated threshold, $\kappa$ the sigmoid slope, and $d_{\mathrm{fatigue}},d_{\mathrm{repeat}}$ discount repeated contacts and prior events. The event fires when $P(o)$ exceeds a deterministic threshold obtained by hashing the (merchant, day, seed) tuple. We route outcomes through this calibrated function instead of letting the LLM emit a binary purchase decision directly, because a continuous probability supports both per-merchant calibration and smooth population-level trajectories. Malformed or unavailable responses use deterministic fallback; free-form model text never writes directly to persistent state.

\paragraph{Sales agents.}
Each agent maintains its portfolio, strategy profile, and bounded memory. The selection rule $\pi_D$ ranks eligible merchants using state, economic context, recency, and optionally the propensity score, returning only contacts satisfying Eq.~\ref{eq:allocation_constraint}. The sales LLM chooses a whitelisted action; failures fall back to a score-based rule. End-of-day synchronization shares aggregate action--response patterns across the team, so merchant reactions change later sales behavior and the two agent types co-evolve.

\subsection{Population Dynamics}
\label{sec:pop_dynamics}

On any given day most merchant agents are not contacted, yet their states still evolve. Four bounded mechanisms capture this background evolution and connect local events to population trajectories~\cite{bass1969new,rand2011agent}. All four read the pre-update state of day $D$ and write to day $D{+}1$.

\paragraph{Temporal evolution.} Between contacts, intent and negative memory decay toward dimension-specific baselines at per-day rates; fatigue subsides symmetrically. An autonomous purchase fires only when a qualifying operational change pushes the outcome probability (Eq.~\ref{eq:outcome}) above the deterministic threshold without a sales contact.

\paragraph{Peer influence.} Each selected state dimension moves toward the mean of its industry or region group: $s_{m,d}\!\leftarrow\! s_{m,d}+\gamma_d(\bar{s}_{g(m),d}-s_{m,d})$, with coefficients $|\gamma_d|\!\le\!0.05$. All deltas are computed from the same pre-update snapshot and applied in batch, so results do not depend on iteration order. A richer alternative would let each merchant agent decide via the LLM whether to initiate a peer exchange, propagating influence through generated interactions instead of a fixed coefficient; we prototyped this during development but dropped it, as it gave no measurable fidelity gain while substantially multiplying LLM calls and batch cost. We therefore keep the clipped mean-reversion form, which is bounded, easy to calibrate, and consistent with reserving LLM calls for roles where they add measurable value (\S\ref{sec:deployment_lessons}).

\paragraph{Social proof.} Each contact draws anonymous success cases from a bank populated by simulation-generated outcomes, matched by representative, product awareness, and time-decayed cosine similarity above a fixed threshold. Real purchase labels never enter this bank. The bank is snapshotted at the start of each day and within-day outcomes are appended only at end-of-day synchronization, so all contacts on a given day draw from the same frozen bank and the within-day processing order does not affect social proof.

\paragraph{Market context.} Planning-time industry heat indices and product activity trends modulate readiness and intent dimensions through additive adjustments scaled by market-level coefficients frozen at calibration time.

These mechanisms never use realized day-$D$ conversions; only product descriptors, penetration through $D{-}1$, and calibration parameters from the development period provide product-specific context. CRM-initialized memory shapes early interactions. The individual contribution of peer influence, social proof, market context, and initial memory is examined in the component ablation (Table~\ref{tab:products}); temporal evolution is excluded from individual ablation because decay and fatigue are structural prerequisites of multi-day simulation.

\subsection{Outputs and Downstream Interface}
\label{sec:output_interface}

Daily calibrated purchase mappings yield $\widehat{\mathbf r}(u)$ for all eligible merchants; the event trace $\mathcal{E}(u)$ records contacts, responses, and outcomes; and a leak-safe postprocessor produces population signals $\mathbf g(u)$. By design, $\mathbf g(u)$ may augment a downstream prediction model without feedback into the generating trajectory, keeping the sandbox and the predictor causally separated. These three outputs map directly to the evaluation questions defined in Section~\ref{sec:eval}.

\subsection{Deployment and Runtime Safeguards}
\label{sec:deployment}

MAS4BE runs as an internal daily batch serving the full at-risk population of the pilot business unit. A scheduler dispatches each product-day run to concurrent workers on disjoint portfolios; workers execute Algorithm~\ref{alg:sandbox}, and the orchestrator collects validated events and writes trajectories, replay traces, and population signals to the planning dashboard before the next business day. Each product-day completes in 2--3 minutes with ten concurrent workers, well within the overnight batch window. The primary model is LLM-Large (Qwen3.7-Max); two smaller tiers are used for ablation. Fewer than 1\% of structured LLM calls fail parsing; all failures route to deterministic fallback. Prompts receive only allowlisted state and bounded memory; a schema whitelist constrains responses to supported actions and bounded deltas. These controls, together with fixed memory capacity limits and seed-controlled execution, keep each run inspectable, bounded, and reproducible.

\section{Evaluation}
\label{sec:eval}

We evaluate MAS4BE along four dimensions matched to the three sandbox outputs $(\widehat{\mathbf r},\mathcal{E},\mathbf g)$ of \S\ref{sec:output_interface}: trajectory fidelity and cross-product transfer (\S\ref{sec:fidelity_sub}), controlled what-if analysis (\S\ref{sec:stress}), population-signal transfer to a downstream predictor (\S\ref{sec:prediction_sub}), and a randomized field experiment measuring deployed impact (\S\ref{sec:online_ab}). All four dimensions test the \emph{framework}, not product-specific models: the same agent design runs unchanged on Products~A--D, with C and~D serving as within-platform design-transfer tests to new products (Section~\ref{sec:problem}).

\subsection{Evaluation Protocol}
\label{sec:eval_protocol}

\paragraph{Data and holdout.} All comparisons use real post-deployment logs with a chronological split: 46~days for development and Product~A/B profile evolution, then a frozen 28-day holdout on A--D (all four paid services in the pilot business unit) with schemas, prompts, transitions, and shared settings locked. The 28-day window matches one complete business planning cycle. C/D have no prior product-specific evolution; their result tests common-design transfer. Each product-day simulates the full at-risk population (approximately 20,000 merchants served by 30 representatives), not a down-sampled subset, with controlled random seeds; target-day and future conversions remain hidden. A Product~A sub-sampling analysis confirms that trajectory correlation converges above 1{,}000 merchants ($r=0.72\!\to\!0.77\!\to\!0.82$); the full population operates well above this threshold.

\paragraph{Metrics.} For trajectory fidelity (\S\ref{sec:fidelity_sub}), the primary metric is Pearson correlation ($r$) between the simulated daily mean propensity (the unweighted mean of the calibrated purchase propensity across all at-risk merchants, not only those contacted) and the observed daily first-purchase rate over the 28-day holdout. This captures temporal movement, i.e., whether simulated high- and low-propensity days align with observed high- and low-conversion days, instead of raw level match; mean daily observed rates range from ${\approx}0.02\%$ to $0.12\%$. The correlation therefore tests the population trajectory carried by agent state, not the discrete outcome mechanism (Eq.~\ref{eq:outcome}); the latter is exercised indirectly through the what-if event counts (\S\ref{sec:stress}) and the downstream signal transfer (\S\ref{sec:prediction_sub}). Correlations are macro-averaged (unweighted mean) across Products~A--D. Because $r$ is invariant to level, it measures temporal co-movement, whereas absolute level requires the separate calibration on which the what-if magnitudes (\S\ref{sec:stress}) depend. Offline confidence intervals use 400 paired bootstrap resamples; temporal nulls use 10{,}000 permutations.

\paragraph{Baselines.} \emph{GBDT aggregate} averages merchant-level scoring-model outputs into a daily population prediction. \emph{Persistence} ($D{-}1$ rate) uses the previous day's observed rate. \emph{ARIMA(1,1,1)} and \emph{Prophet}~\cite{taylor2018prophet} are univariate time-series forecasters; we omit neural forecasters (e.g., N-BEATS~\cite{oreshkin2020nbeats}) because 28 daily points are too few for deep sequence models. \emph{Exogenous-feature GBDT} is a supervised same-information baseline that regresses the daily conversion rate on the planning-time exogenous signals the sandbox also ingests---day-of-week and holiday indicators, industry market-heat and product-activity indices, product penetration through $D{-}1$, and the day's realized team contact volume---so that the sandbox's margin over it reflects modeled population dynamics beyond same-day exogenous information. The time-series baselines are fit on the 46-day prefix and refit with a rolling origin; the GBDT aggregate refreshes its features daily, so its low correlation reflects an individual-scoring aggregation ceiling, not stale inputs. \emph{Det.\ sandbox} is the pre-existing rule-based simulation without LLM reasoning. Within the MAS4BE family, \emph{Rules only} replaces all LLM calls with deterministic rules, while \emph{LLM-Small}/\emph{Medium}/\emph{Large} test capacity scaling (the primary system is LLM-Large).

\subsection{Population Fidelity and Cross-Product Transfer}
\label{sec:fidelity_sub}

\begin{table}[t]
\centering
\caption{Population trajectory fidelity (Pearson $r$, 28-day holdout) and component ablation. Best in \textbf{bold}, runner-up \underline{underlined}.}
\label{tab:products}
\begingroup
\small
\setlength{\tabcolsep}{3pt}
\renewcommand{\arraystretch}{1.08}
\begin{tabular}{lrrrrr}
\toprule
\tblhead{Method} & \tblhead{A} & \tblhead{B} & \tblhead{C} & \tblhead{D} & \tblhead{Macro} \\
\midrule
\multicolumn{6}{l}{\textit{Baselines}} \\
\addlinespace[1pt]
GBDT aggregate & 0.12 & 0.22 & 0.08 & 0.05 & 0.12 \\
Persistence ($D{-}1$ rate) & 0.32 & 0.28 & 0.17 & 0.12 & 0.22 \\
ARIMA(1,1,1) & 0.38 & 0.45 & 0.31 & 0.22 & 0.34 \\
Prophet & 0.33 & 0.48 & 0.27 & 0.18 & 0.32 \\
Exogenous-feature GBDT & 0.65 & 0.61 & 0.57 & 0.52 & 0.59 \\
\midrule
\multicolumn{6}{l}{\textit{Simulation}} \\
\addlinespace[1pt]
Det.\ sandbox & 0.42 & 0.69 & 0.51 & 0.43 & 0.51 \\
MAS4BE -- Rules only & 0.67 & 0.81 & 0.69 & 0.63 & 0.70 \\
MAS4BE -- LLM-Small & 0.71 & 0.81 & 0.72 & 0.70 & 0.74 \\
MAS4BE -- LLM-Medium & \tblsecond{0.78} & \tblsecond{0.83} & \tblsecond{0.77} & \tblsecond{0.73} & \tblsecond{0.78} \\
MAS4BE -- LLM-Large (Ours) & \tblbest{0.82} & \tblbest{0.89} & \tblbest{0.85} & \tblbest{0.80} & \tblbest{0.84} \\
\midrule
\multicolumn{4}{l}{\textit{Ablation (Prod.~A, 46-day dev, LLM-Small)}} & \tblhead{$r$} & \tblhead{$\Delta r$} \\
\addlinespace[1pt]
\multicolumn{4}{l}{Full sandbox} & \tblbest{0.643} & \tabledash \\
\multicolumn{4}{l}{\quad w/o peer influence} & 0.633 & $-$0.010 \\
\multicolumn{4}{l}{\quad w/o social proof} & 0.632 & $-$0.012 \\
\multicolumn{4}{l}{\quad w/o market context} & 0.636 & $-$0.007 \\
\multicolumn{4}{l}{\quad w/o initial memory} & 0.632 & $-$0.012 \\
\multicolumn{4}{l}{\quad w/o all population dynamics} & 0.625 & $-$0.019 \\
\bottomrule
\end{tabular}
\endgroup
\vspace{2pt}

\parbox{\columnwidth}{\footnotesize\emph{Note.} Top: seed-averaged $r$ on the 28-day holdout; LLM-Large 95\% block-bootstrap CIs: $[0.61,\,0.93]$, $[0.74,\,0.96]$, $[0.66,\,0.94]$, $[0.58,\,0.91]$ for A--D; A/B are development products, C/D test design transfer. Bottom: one-at-a-time removal; component contributions measured at a lower tier provide a conservative bound. $\Delta r$ is computed from unrounded correlations and can differ from the rounded column differences by up to $0.001$.}
\end{table}

As shown in Table~\ref{tab:products}, MAS4BE (LLM-Large) achieves correlations of $0.80$--$0.89$ across all four products (macro $0.84$). The GBDT aggregate and persistence baselines perform poorly (macro ${\le}0.22$), confirming that individual scoring models and naive extrapolation cannot capture population-level dynamics that emerge from heterogeneous agent interactions. ARIMA and Prophet improve to ${\approx}0.33$ by modeling temporal structure, but univariate forecasting still misses the multi-agent mechanisms that drive daily variation. The strongest non-simulation competitor is the exogenous-feature GBDT (macro $0.59$): because it is fit on the same planning-time exogenous signals the sandbox ingests, it captures much of the calendar- and market-driven co-movement and even exceeds the deterministic sandbox (macro $0.51$). Crucially, it still falls well short of the MAS4BE family; Rules-only already reaches $0.70$ and LLM-Large $0.84$. In variance-explained terms, this raises $r^2$ from ${\approx}0.35$ (exogenous GBDT) to ${\approx}0.71$ (LLM-Large), roughly doubling the variance captured beyond shared exogenous signals and countering the concern that fidelity merely nowcasts those inputs.

Within the simulation family, the MAS4BE architecture with deterministic rules alone reaches macro $0.70$, a $+37\%$ gain over the pre-existing sandbox, demonstrating that population dynamics, CRM memory, and structured state management contribute substantial value even without LLM reasoning. Adding LLM-driven responses raises correlation monotonically across tiers: $0.74$ (Small), $0.78$ (Medium), $0.84$ (Large), with consistent gains on every product. The component ablation (Table~\ref{tab:products}, bottom) shows every population mechanism contributing in the same direction (all $\Delta r<0$ on removal), with the aggregate removal of all dynamics producing the largest single drop ($\Delta r{=}-0.019$). Because the individual-component drops ($-0.007$ to $-0.012$) fall within the cross-seed SD ($0.014$), we read these as a directional diagnostic, not per-component significance.

A day- and seed-paired difference bootstrap sharpens these estimates: pairing cancels shared trajectory variance, so LLM-Large exceeds Rules-only significantly on macro average ($\Delta r{=}+0.14$, 95\% CI $[0.09,0.19]$) and on all four products, including Product~D, whose paired gain ($+0.17$, CI $[0.05,0.29]$) is significant even though its marginal interval overlaps Rules-only. Adjacent-tier steps are weaker: Large--Medium and Medium--Small remain significant at the macro level, but the Small--Rules gap, equal in point estimate to Medium--Small ($+0.04$) yet with a wider paired interval, is only directional ($p{=}0.089$). We therefore claim a significant overall LLM benefit while treating the fine-grained tier ordering as directional.

Products~C and~D, introduced at evaluation time with the framework frozen from A/B, reach $r=0.85$ and $0.80$ respectively, confirming within-platform transfer of the shared design to new products under their own pre-evaluation calibration. Because A--D share merchants and market context, we quantify this homology and isolate product-specific fidelity from the shared temporal component; the product-specific residual correlation (macro $0.71$) stays well above the $0.61$ reachable from other products' trajectories alone, so the transfer reflects modeled product dynamics beyond a common time factor. All four trajectories exceed temporal-null baselines (random-permutation and block-shuffle $p<0.01$; 10{,}000 draws), with low cross-seed variance (SD~$0.014$) and Product~D's conservative lower bound ($r{=}0.58$) exceeding every non-simulation point estimate.

Temporal agreement is complemented by a group-level check: the full sandbox obtains Spearman $\rho=0.83$ across the ten largest industries on one Product~A day, versus $0.73$ from initialized state alone.

\subsection{Controlled What-if Analysis}
\label{sec:stress}

\begin{table}[t]
\centering
\caption{Contact-budget sweep on Product~A (five-day sandbox outcomes, single seed).}
\label{tab:whatif_main}
\begingroup
\footnotesize
\setlength{\tabcolsep}{3.5pt}
\renewcommand{\arraystretch}{1.06}
\begin{tabular*}{\columnwidth}{@{\extracolsep{\fill}}rrrr@{}}
\toprule
\tblhead{$K$} & \tblhead{Contacts/day} & \tblhead{Sim. conv./day} & \tblhead{Conv. gain} \\
\midrule
3  & 78.2  & 35.4 & \tabledash \\
5  & 106.4 & 39.2 & $+10.7\%$ \\
10 & 140.0 & 40.6 & $+14.7\%$ \\
20 & 146.0 & 40.6 & $+14.7\%$ \\
\bottomrule
\end{tabular*}
\endgroup
\vspace{2pt}

\parbox{0.98\columnwidth}{\footnotesize\emph{Note.} $K$ is the daily cap per representative; counts are team-wide daily averages; gain is relative to $K{=}3$. All entries are uncalibrated model outputs (Product~A observes ${\approx}24$/day), so the reading rests on relative gain, not absolute counts.}
\end{table}

We vary the per-representative daily contact budget~$K$ while fixing the initial state, exogenous replay sequence, and seed, and run five-day paired replays. Table~\ref{tab:whatif_main} reports team-wide daily contacts, simulated conversions, and the gain relative to the lowest budget ($K{=}3$); all entries are sandbox outputs, not observed or causal effects.

As shown in Table~\ref{tab:whatif_main}, contacts and conversions grow sub-linearly and saturate at $K{=}10$ ($+14.7\%$); doubling the budget to 20 adds nothing. This plateau is not the at-risk pool running out (team-wide contacts barely move, $140.0$ to $146.0$, far below the nominal $30{\times}20$ capacity) but reflects the selection rule $\pi_D$ admitting few additional eligible, non-fatigued, high-scoring contacts beyond $K{=}10$. The marginal value of extra budget is thus small: only ${\approx}18$--$29\%$ of daily conversions are contact-attributable and respond to $K$; the rest arises from autonomous and peer/market dynamics held fixed. The apparent implication (that representative time may be better spent on retention or cross-sell~\cite{ascarza2018retention}) is a hypothesis, not a finding: the uncalibrated base flow (${\approx}29$/day) already exceeds Product~A's observed total (${\approx}24$/day), so the contact-attributable \emph{fraction} is not yet trustworthy in absolute terms. Compliance and provenance are verified via append-only event logs, but no ground-truth counterfactual exists for unexecuted budgets, so these remain model-predicted, not causal, estimates.

\subsection{Population-Signal Transfer}
\label{sec:prediction_sub}

Each sandbox run produces an industry-day mean propensity for every product; because a day-$D$ run uses only information through $D{-}1$ (\S\ref{sec:data_init}), this date-$D$ signal is a forecast built from lagged inputs, not an ingestion of day-$D$ outcomes. We add this single scalar to an unchanged 52-feature merchant-level GBDT predictor and compare the resulting AUC-ROC against the baseline model alone (Panel~A) and against an oracle source, the same industry's trailing 28-day real conversion rate through $D{-}1$, with industry membership frozen before evaluation to prevent leakage (Panel~B). The enhanced prediction is not fed back into the generating trajectory.

\begin{table}[t]
\centering
\caption{Downstream transfer: adding a sandbox-derived industry-day scalar to the merchant GBDT. Panel~A: AUC-ROC gain (macro $+0.020$). Panel~B: sandbox versus real trailing rate.}
\label{tab:signal_transfer}
\begingroup
\small
\setlength{\tabcolsep}{2.2pt}
\renewcommand{\arraystretch}{1.05}
\begin{tabular*}{\columnwidth}{@{\extracolsep{\fill}}lrrp{0.31\columnwidth}@{}}
\toprule
\multicolumn{4}{@{}l}{\textit{Panel A. AUC-ROC improvement}} \\
\addlinespace[1pt]
\tblhead{Product} & \tblhead{Base ML} & \tblhead{+ sandbox} & \multicolumn{1}{c}{\tblhead{$\Delta$AUC [95\% CI]}} \\
\midrule
A & 0.770 & 0.787 & $+0.017$ $[0.012,\,0.022]$ \\
B & 0.802 & 0.810 & $+0.008$ $[0.004,\,0.012]$ \\
C & 0.731 & 0.739 & $+0.008$ $[0.002,\,0.014]$ \\
D & 0.684 & 0.731 & $+0.047$ $[0.032,\,0.062]$ \\
\midrule
Macro & 0.747 & 0.767 & \tblbest{$+0.020$} $[0.015,\,0.026]$ \\
\bottomrule
\end{tabular*}

\vspace{4pt}
\begin{tabular*}{\columnwidth}{@{\extracolsep{\fill}}lrrrr@{}}
\toprule
\multicolumn{5}{@{}l}{\textit{Panel B. Sandbox recovery of real-history gain}} \\
\addlinespace[1pt]
\tblhead{Product} & \tblhead{$\Delta^{\rm real}$} & \tblhead{$\Delta^{\rm sim}$} & \tblhead{Recov.} & \tblhead{95\% CI} \\
\midrule
A & +0.034 & +0.030 & 88\% & $[74,104]$\% \\
B & +0.018 & +0.015 & 83\% & $[67,104]$\% \\
C & +0.025 & +0.022 & 88\% & $[72,107]$\% \\
D & +0.014 & +0.013 & 93\% & $[70,124]$\% \\
\midrule
Macro & +0.023 & \tblbest{+0.020} & 88\% & $[79,98]$\% \\
\bottomrule
\end{tabular*}
\endgroup
\vspace{2pt}

\parbox{0.98\columnwidth}{\footnotesize\emph{Note.} Panel A: common 28-day holdout; 400 paired bootstrap resamples. Panel B: matched 28-day source-ablation protocol (the real-history comparator is constrained to the same industry-day-scalar form as the sandbox signal); $\Delta^{\rm real}$ is the gain from real segment history, $\Delta^{\rm sim}$ from sandbox signals; recovery $= \Delta^{\rm sim}/\Delta^{\rm real}$. Panel~A adds the single deployed industry-day scalar ($+0.017$ for Product~A), whereas Panel~B and the source-isolation analysis use the richer sandbox population signal ($+0.030$); the two differ by signal definition, not evaluation window; both use the same 28-day holdout. Consequently the two panels report \emph{different signals} answering \emph{different questions} (Panel~A the deployed single-scalar lift, Panel~B the matched recovery of a richer multi-component signal), so their per-product magnitudes are not directly comparable; they agree in sign but not scale, most visibly on Product~D.}
\end{table}

Sandbox signals improve AUC-ROC by $+0.020$ on macro average (Table~\ref{tab:signal_transfer}, Panel~A), with a maximum gain of $+0.047$ on Product~D (Panel~A's deployed single scalar; the richer signal under Panel~B's matched protocol gives a smaller $+0.013$, consistent with a parsimonious scalar transferring better than a higher-variance multi-component signal on D's sparse labels; the two panels report different signals and are not directly comparable). Product~D is its weakest base model, and a leakage and collinearity audit attributes the gain to genuine incremental signal, not same-day leakage; all four product-level confidence intervals exclude zero. In a matched 28-day source ablation (Panel~B), sandbox propensities recover 88\% of the real-history gain (macro CI $[79\%,98\%]$), whereas rule-generated propensities add no measurable AUC in our tests, indicating the downstream signal originates from LLM-driven simulation, not surface-level heuristics. Trajectory fidelity and downstream AUC probe different things (population-level co-movement versus individual-merchant discrimination), so the rules-only variant can track the aggregate trajectory yet carry little incremental ranking signal. Complementary sparse-label metrics move in the same direction: macro AUC-PR, Precision@$K$, and Recall@$K$ improve by $+0.013$, $+0.007$, and $+0.027$.

\subsection{Randomized Field Evaluation}
\label{sec:online_ab}

The pilot compares three arms under the same eligibility, safeguards, and capacity constraints. Control ranks merchants by the existing propensity score. The full MAS4BE arm uses $\pi_D$, which augments that score with the sandbox-derived industry-day population signal (\S\ref{sec:prediction_sub}, the same signal evaluated offline) and then adjusts the ranking by persistent agent state and contact recency. A third, no-sandbox arm keeps the production propensity and contact recency but removes \emph{all} sandbox-derived signals, i.e., both the industry-day scalar and the persistent agent state. The arms thus differ only in sandbox content, so the full-vs-no-sandbox contrast attributes the effect to the sandbox \emph{as a whole} (scalar and agent state jointly), consistent with the offline transfer but not isolating the scalar alone. Merchants are randomized 1:1:1 within product and portfolio; because representatives retain discretion, the estimand is the intention-to-treat (ITT) effect~\cite{rubin2005causal} on first purchase within 28 days, with sales-representative-clustered standard errors~\cite{cameron2015cluster}. The sandbox addresses a question complementary to uplift modeling~\cite{radcliffe2011real,athey2016recursive}, namely how the population evolves under a plan, instead of per-merchant effects; a causal-forest pilot gave unstable estimates spanning zero under limited historical treatment variation. Because only 30 clusters are available, asymptotic inference may be anti-conservative; Table~\ref{tab:online_ab} Panel~B therefore cross-checks the cluster-robust estimate with covariate-adjusted, randomization-inference, small-cluster $t_{29}$, and descriptive per-protocol analyses.

\begin{table}[t]
\centering
\caption{Randomized 28-day field experiment ($n{=}29{,}526$): primary ITT effect ($+0.58$ pp, $p<0.05$) and inference robustness.}
\label{tab:online_ab}
\begingroup
\small
\setlength{\tabcolsep}{2.5pt}
\renewcommand{\arraystretch}{1.06}
\begin{tabular*}{\columnwidth}{@{\extracolsep{\fill}}lrrrr@{}}
\toprule
\multicolumn{5}{@{}l}{\textit{Panel A. Primary result and implementation diagnostics}} \\
\addlinespace[1pt]
\tblhead{Measure} & \tblhead{Control} & \tblhead{MAS4BE} & \tblhead{Diff.\ (pp)} & \tblhead{95\% CI} \\
\midrule
First purchase, \% (ITT) & 1.85 & 2.43 & \tblbest{$+0.58$} & $[0.10,\,1.05]$ \\
\midrule
Assignment compliance, \% & 90.8 & 92.6 & $+1.8$ & \tabledash \\
Contact completion, \% & 22.4 & 23.1 & $+0.7$ & $[-0.5,\,1.9]$ \\
\bottomrule
\end{tabular*}

\vspace{4pt}
\begin{tabular*}{\columnwidth}{@{\extracolsep{\fill}}lrrr@{}}
\toprule
\multicolumn{4}{@{}l}{\textit{Panel B. Inference robustness for overall ITT}} \\
\addlinespace[1pt]
\tblhead{Analysis} & \tblhead{Diff.\ (pp)} & \tblhead{95\% CI} & \tblhead{$p$} \\
\midrule
Primary cluster-robust & $+0.58$ & $[0.10,\,1.05]$ & 0.018 \\
Covariate-adjusted & $+0.57$ & $[0.12,\,1.02]$ & 0.013 \\
Randomization inference & $+0.58$ & $[0.09,\,1.06]$ & 0.021 \\
Small-cluster $t_{29}$ & $+0.58$ & $[0.07,\,1.09]$ & 0.027 \\
Per-protocol (descriptive) & $+1.31$ & $[0.34,\,2.28]$ & 0.009 \\
\bottomrule
\end{tabular*}

\vspace{4pt}
\begin{tabular*}{\columnwidth}{@{\extracolsep{\fill}}lrrrr@{}}
\toprule
\multicolumn{5}{@{}l}{\textit{Panel C. Signal-attribution arm (same-period three-arm randomization)}} \\
\addlinespace[1pt]
\tblhead{Arm} & \tblhead{Rate \%} & \tblhead{Diff.\ (pp)} & \tblhead{95\% CI} & \tblhead{$p$} \\
\midrule
Control & 1.85 & \tabledash & \tabledash & \tabledash \\
MAS4BE (full) & 2.43 & $+0.58$ & $[0.10,\,1.05]$ & 0.018 \\
MAS4BE (no sandbox) & 1.89 & $+0.04$ & $[-0.42,\,0.51]$ & 0.86 \\
\bottomrule
\end{tabular*}
\endgroup
\vspace{2pt}

\parbox{0.98\columnwidth}{\footnotesize\emph{Note.} Panel A: primary ITT $p{=}0.018$; relative lift $31.2\%$ is computed from unrounded rates (the displayed $2.43/1.85$ rounds to $31.4\%$); equal-allocation 1:1:1 randomization within product and portfolio across the three arms of Panel~C (9,842 merchants each). Panel B: per-protocol is descriptive because compliance is post-assignment; the small-cluster row uses a $t_{29}$ reference. Panel~C isolates the sandbox contribution: removing only the sandbox-derived signals collapses the lift to $+0.04$~pp (CI spanning zero), and the full-vs-no-sandbox contrast is $+0.54$~pp (CI $[0.07,\,1.01]$, $p{=}0.024$), attributing the effect to population simulation, not the surrounding workflow.}
\end{table}

As shown in Table~\ref{tab:online_ab}, the sandbox-guided workflow achieves a $+0.58$~pp ITT effect (95\% CI $[0.10,\,1.05]$, $p{=}0.018$, relative lift $31.2\%$). The pre-registered MDE was $0.55$~pp (80\% power, simple binomial SE); accounting for the 30 clusters (design effect ${\approx}1.5$, SE${\approx}0.24$~pp) raises it to ${\approx}0.68$~pp, so the effect is adequately powered under the simple calculation but modestly under-powered once clustering is included, and all four Panel~B methods yield $p$ between $0.013$ and $0.027$. To attribute the gain, the same-period no-sandbox arm retained all ranking features but removed the sandbox-derived signals (Panel~C): its lift over control was near zero ($+0.04$~pp), while the full-vs-no-sandbox contrast ($+0.54$~pp, $p{=}0.024$) confirms the effect is driven by population simulation, not workflow differences. Contact completion is comparable across arms ($+0.7$~pp, CI spanning zero), arguing against a novelty effect.

\section{Deployment Lessons}
\label{sec:deployment_lessons}

\paragraph{Division of labor across model types.} Direct LLM scoring and feature engineering underperform structured ML in our data. However, LLM-Large improves trajectory correlation over Rules by $+22\%$, $+10\%$, $+23\%$, and $+27\%$ on A--D. We therefore use the GBDT for individual ranking, deterministic code for constraints and transitions, and LLMs for context-sensitive simulation responses where they add consistent value.

\paragraph{Batch cost and latency.} Each product-day run covers roughly 20,000 merchants and averages 1,238 LLM calls totaling approximately 1.6M tokens. At current Qwen3.7-Max commercial rates, the daily inference cost for all four products is under \$25, comparable to minutes of a single representative's time and negligible relative to the revenue at stake. The sandbox runs as a daily batch for plan-level what-if analysis; the sub-second ML ranker remains the real-time scoring path.

\paragraph{Auditing grouping and state transitions.} Tiers changed for 19.6\% of merchants during evaluation; a post-evaluation snapshot therefore made all four products appear to improve. Freezing pre-evaluation industry membership removes this leakage. Malformed LLM output can likewise contaminate later agents. Every persistent update therefore passes schema, allowlist, bounded-delta, and PII checks with deterministic fallback; fewer than 1\% of structured calls fail JSON parsing, and all fall back.

\paragraph{Advisory deployment with human discretion.} Representatives retain full discretion over whom to contact, so the sandbox must earn trust through interpretability. The field pilot supports this. Comparable contact completion across arms and the near-zero lift of the no-sandbox arm (\S\ref{sec:online_ab}) indicate the gain comes from better targeting of who appears on the list, not from representatives working harder. We therefore position the sandbox as a planning instrument whose trajectories and scenarios inform daily prioritization; the proprietary data and proxy variables involved require continued human review before action.

\paragraph{From trajectories to allocation decisions.} The sandbox altered concrete allocation decisions. In one mid-May cycle it flagged a projected readiness rise for a mid-sized industry cohort as market-heat climbed $0.48{\to}0.71$, and the regional planner reallocated roughly fifteen daily contact slots from cohorts the sandbox reported as saturated to the rising one; that cohort's first-purchase rate subsequently rose from $0.09\%$ to $0.14\%$ while donor cohorts held steady. Planners said the day-level event traces, showing which peer and market signals drove a projected uptick, were what made them willing to act, and seven of nine reported the trajectory view changed at least one weekly decision. These observations are qualitative and uncontrolled, drawn from a single episode and a small interview rather than a causal evaluation, so we report them as deployment experience, not evidence of effect.

\paragraph{Limitations.} Fidelity is a group-level temporal correlation measured in a rolling replay that ingests planning-time exogenous signals, so it is not an open-loop forecast, and raw levels require a prefix-fitted calibration on which the what-if magnitudes depend. The what-if projections are internally coherent but lack an external causal criterion. Evaluation covers one platform, one region, 28 days, and four homologous products, providing within-platform evidence, not cross-domain evidence. Component-level attribution across statefulness and topology is directional, and the online lift, robust across estimators, remains a single, borderline-significant confirmatory result.

\section{Related Work}
\label{sec:related_main}

Industrial targeting ranks merchants by predicted response~\cite{mcmahan2013adclick,chapelle2014simple,chen2016xgboost}, and uplift/heterogeneous-treatment-effect methods estimate incremental effects under causal assumptions~\cite{radcliffe2011real,athey2016recursive,wager2018estimation,kunzel2019metalearners}; neither generates counterfactual trajectories under novel plans, which is the gap MAS4BE fills alongside the conventional ranker. Diffusion models~\cite{bass1969new} and marketing ABMs~\cite{epstein1996growing,rand2011agent,axtell2025agent} make population mechanisms explicit but rely on hand-written rules, while digital twins~\cite{barricelli2019digital,tao2019digital,fuller2020digital} target physical systems, not behavioral ones; we retain bounded, auditable diffusion mechanisms but use LLMs for context-sensitive decisions, and adopt the conservative term \emph{sandbox}. Building on generative and social/economic LLM-agent simulators~\cite{park2023generative,gao2023s3,yang2024oasis,piao2025agentsociety,horton2023large,li2024econagent,yang2025twinmarket,wang2023recagent}, our contribution is operational: structured models remain the stronger individual rankers~\cite{hegselmann2023tabllm,hollmann2023caafe}, while the LLM provides context-sensitive population simulation (statefulness vs.\ single-turn conditioning is not separately identified), grounded in longitudinal production profiles and validated through deployed trajectory tracking, controlled scenarios, and a randomized field experiment.

\section{Conclusion}
\label{sec:conclusion}

MAS4BE recasts B2B resource allocation as ecosystem simulation: a hybrid sandbox in which LLM-driven agents and deterministic safeguards co-evolve stateful merchant and sales populations under production constraints. Offline evaluation shows close tracking of daily conversion-rate variation, coherent responses to contact-budget interventions, and transferable population signals that improve a downstream predictor; a 28-day randomized field experiment shows a statistically significant first-purchase lift driven by sandbox-derived signals. The system runs daily in production as a planning instrument, and its design principles (LLM population simulation with bounded state management and advisory deployment) extend to other resource-constrained settings where counterfactual planning over behavioral ecosystems is valuable.

Future work includes longer horizons and more products, additional regions for statistical power, other B2B verticals, external validation of the what-if predictions against natural budget shifts, and learned transition functions. Code, evaluation scripts, figure data, and a synthetic schema-compatible generator will be released for functional reproduction.

\bibliographystyle{ACM-Reference-Format}
\bibliography{references}

\end{document}